\documentclass[conference]{IEEEtran}
\IEEEoverridecommandlockouts

\usepackage{times}

\usepackage[numbers]{natbib}
\usepackage{multicol}
\usepackage[bookmarks=true]{hyperref}
\usepackage{xcolor}
\usepackage{float}
\usepackage{graphicx}

\begin{document}

\title{VLA-3D: A Dataset for 3D Semantic Scene Understanding and Navigation}

\IEEEaftertitletext{\vspace{-1em}}
\author{\authorblockN{Haochen Zhang, Nader Zantout, Pujith Kachana, Zongyuan Wu, Ji Zhang, Wenshan Wang}
\authorblockA{Robotics Institute, Carnegie Mellon University \\
\{haochen4, nzantout, pkachana, zongyuaw, zhangji, wenshanw\}@andrew.cmu.edu}%
\thanks{Presented at the 1st Workshop on Semantic Reasoning and Goal Understanding in Robotics (SemRob), Robotics Science and Systems (RSS 2024)}%
\vspace{-1em}
}

\maketitle

\begin{abstract}

With the recent rise of Large Language Models (LLMs), Vision-Language Models (VLMs), and other general foundation models, there is growing potential for multimodal, multi-task embodied agents that can operate in diverse environments given only natural language as input. One such application area is indoor navigation using natural language instructions. However, despite recent progress, this problem remains challenging due to the spatial reasoning and semantic understanding required, particularly in arbitrary scenes that may contain many objects belonging to fine-grained classes. To address this challenge, we curate the largest real-world dataset for Vision and Language-guided Action in 3D Scenes (VLA-3D), consisting of over 11.5K scanned 3D indoor rooms from existing datasets, 23.5M heuristically generated semantic relations between objects, and 9.7M synthetically generated referential statements. Our dataset consists of processed 3D point clouds, semantic object and room annotations, scene graphs, navigable free space annotations, and referential language statements that specifically focus on view-independent spatial relations for disambiguating objects. The goal of these features is to specifically aid the downstream task of navigation, especially on real-world systems where some level of robustness must be guaranteed in an open world of changing scenes and imperfect language. We also aim for this dataset to aid the development of interactive agents that can both respond to commands and ask and answer questions regarding a scene. We benchmark our dataset with current state-of-the-art models to obtain a performance baseline. All code to generate and visualize the dataset is publicly released \footnote{https://github.com/HaochenZ11/VLA-3D}. With the release of this dataset, we hope to provide a resource for progress in semantic 3D scene understanding that is robust to changes and one which will aid the development of interactive indoor navigation systems.

\end{abstract}

\IEEEpeerreviewmaketitle

\section{Introduction}
Methods combining vision and language have been evolving rapidly with the advent of both Large Language Models (LLMs) \cite{achiam2023gpt, touvron2023llama, team2023gemini} and Vision-Language Models (VLMs) \cite{radford2021learning, ramesh2021zero, liu2024visual} pre-trained on significant amounts of data, tackling various 2D tasks such as Visual Question Answering (VQA) \cite{antol2015vqa}, image retrieval \cite{karpathy2014deep}, and image captioning \cite{radford2021learning}. As we progress towards generalizable embodied intelligence, there is a need for methods that are capable of reasoning in 3D-space and interacting with humans. Using natural language for example, humans are able to refer to objects in a 3D scene in a way that disambiguates the target object, often using the utterance of ``least effort" \cite{zipf2016human} and making use of relative spatial relationships. An agent that can similarly solve such a problem would be particularly valuable in robotics fields such as indoor-navigation with applications as in-home assistants.

The pursuit of such agents that can identify and understand 3D scenes, consolidate visual input with language semantics, and display robust performance for real-world deployment, however, presents various challenges. First, the scene can have hundreds of objects, contain objects belonging to fine-grained classes, and have many similar objects \cite{ramakrishnan2021habitat}. Second, human referential language often involves spatial reasoning, affordances, open-vocabulary language, and may even be incorrect or refer to something that does not exist, e.g. \textit{``the remote on the table"} when the remote is actually on the sofa. Third, the scale of available vision-language data in the 3D space pales in comparison to the amount of 2D data, which was crucial to the success of 2D vision-language learning methods \cite{krishna2017visual, changpinyo2021conceptual}. Despite impressive recent advancements with foundation models, such problems remain difficult when applied to robotics as current methods fail to offer the accuracy and robustness needed for real-world deployment \cite{huang2022multi}.

To this end, we propose a novel dataset based on 3D scenes from a diverse set of existing scans of indoor environments that provides a unique resource for training referential object grounding methods. Building on top of the scans, we provide 1) point clouds as they enable learning directly from 3D geometric and visual information \cite{armeni20163d}, 2) extracted object-level attributes and semantic class labels for discriminating and categorizing objects, 3) large-scale scene graphs with spatial relations for a structured representation of a scene, 4) referential language statements to support vision-language grounding using natural language, and 5) traversable free space annotations to explicitly connect to downstream navigation tasks. The inclusion of dense scene graphs and traversable free space are two features that particularly distinguish our dataset from previous object-referential datasets. The scene graphs allow for a robust representation for the semantics in each scene that can be used to guide the grounding task and also to infer when the language statement is invalid. Free space annotations give the ability to generate referential statements that refer not just to objects but to spaces or paths.  

\begin{figure}
\centering
\includegraphics[width=0.4\textwidth]{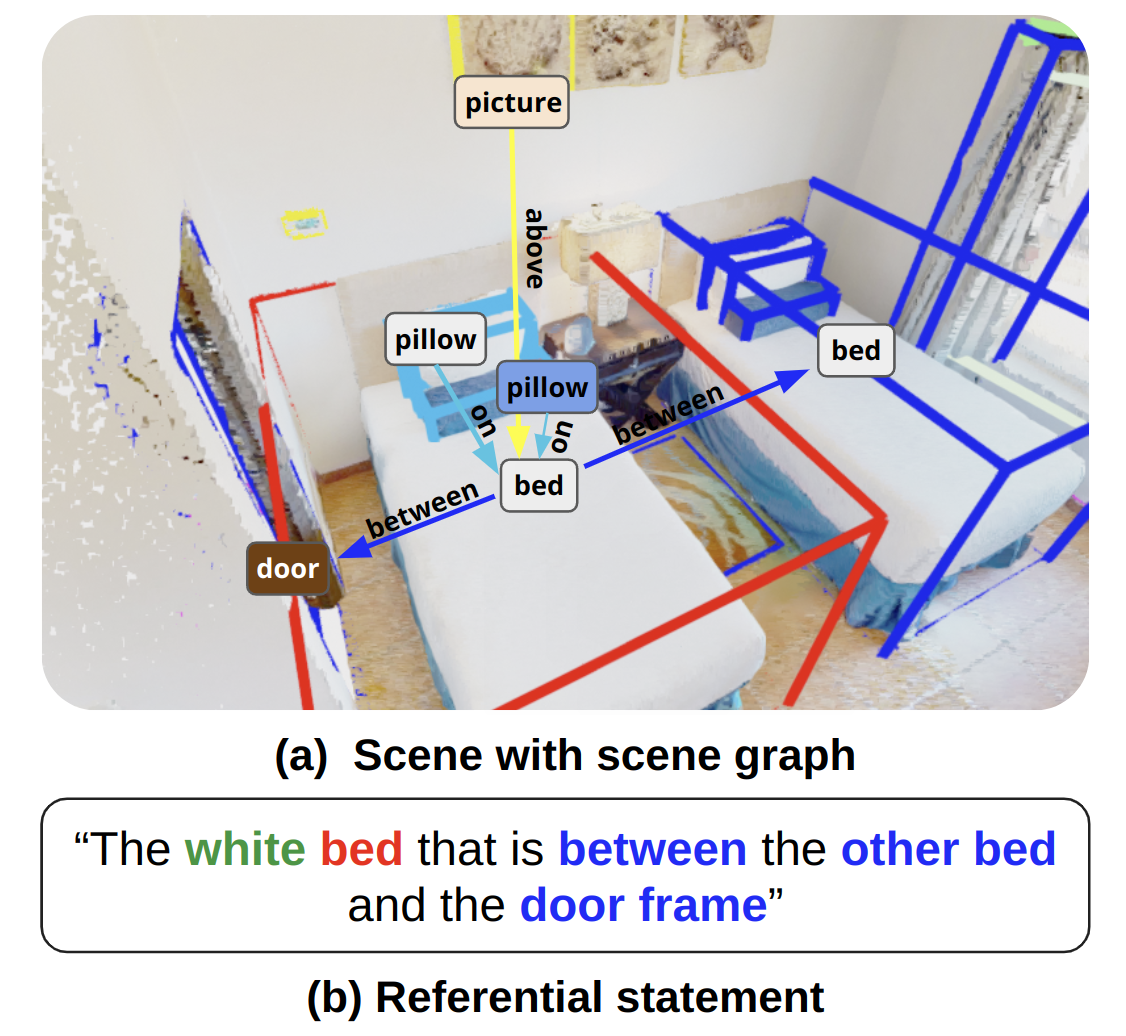}
\caption{Sample region from the dataset visualized with (a) a scene graph (a) and (b) a corresponding referential statement}. 
\label{fig:data_sample}
\end{figure}

Along with our dataset, we also release the code for the entire dataset generation process, demonstrating that synthetic heuristic-based generation methods can aid the efficient generation of large-scale datasets. A custom dataset visualizer tool is also provided to visualize individual scenes and regions from our dataset. With our dataset, we test two state-of-the-art (SOTA) referential object grounding baseline models on our data to verify that such low-level semantic understanding remains a challenging problem and provide a starting point to the identification of where SOTA methods may fail. This can then aid the subsequent development of higher-fidelity 3D vision-language methods that reason over real-world scenes. A sample from our dataset is shown in Figure \ref{fig:data_sample}.

\section{Related Work}

\subsection*{Object Referential Datasets}
The referential object-grounding task has been defined and explored in datasets such as CLEVR \cite{johnson2017clevr} in the 2D space and ReferIt3D \cite{achlioptas2020referit3d}, ScanRefer \cite{chen2020scanrefer}, SceneVerse \cite{jia2024sceneverse} in the 3D space. While these datasets are similar in the style of their referential statements, the statements are often unintuitive and unnatural compared to human referential statements. E.g, using clock bearings to describe spatial location or using many redundant or subjective descriptors such as ``\textit{comfy}" \cite{achlioptas2020referit3d}. Both ReferIt3D and ScanRefer are also of a smaller scale and focus only on a single scene data source in which all scenes are single-room, making them less suitable for downstream navigation tasks. SceneVerse scales the data up by curating a much larger dataset and generating statements synthetically, then using an LLM for rephrasing. Despite the increase in scale, the LLM-rephrased statements are often unnatural (e.g. \textit{``the chair stands proudly against the wall"}), and the templates lack explicit references to attributes like size, color, and shape which humans often use for object reference. As a result, models trained on SceneVerse still performed poorly on the Nr3D benchmark \cite{jia2024sceneverse}.

\subsection*{Semantic Scene Graph Datasets}
Generating scene graphs from 3D scenes has also been explored in 3DSSG \cite{wald2020learning}, Hydra \cite{hughes2022hydra}, HOV-SG \cite{werby2024hierarchical}, and ConceptGraphs \cite{gu2023conceptgraphs}. 3DSSG focuses on predicting scene graphs automatically, resulting in generated graphs that can miss relations or generate redundant ones. The main use case is scene retrieval from a set of scenes which is different from the navigation paradigm where relations must help disambiguate objects or locations within a single scene. 
In Hydra, a system is developed to build 3D scene graphs in real-time but does not include explicit language-grounding. While HOV-SG and ConceptGraphs both build open-vocabulary scene graphs, the language-guided navigation task they're designed for involves referring to an object mainly using region references rather than fine-grained inter-object relations.
\subsection*{Instruction-Following Datasets}
 Multiple works have also explored language-guided navigation through instruction-following statements, often specifying a series of steps to move between regions in a large scene. Common datasets include Room Across Room \cite{ku2020room} and Room-2-Room \cite{mattersim}. While these datasets involve spatial and directional references, the task is different from ours as it focuses on a series of distinct steps between rooms rather than explicit fine-grained inter-object spatial relations. The instruction-following task not only requires knowledge of what objects are present, but also exactly where objects and rooms are relative to each other in a multi-room scene, making it difficult to be robust to scene changes or imperfect language.
\subsection*{Referential Object Grounding}
A number of papers have explored the task of learning referential object grounding, mainly on either the ReferIt3D benchmark or the ScanRefer task. These include BUTD-DETR \cite{jain2022bottom}, MVT \cite{huang2022multi}, ViL3DRel \cite{chen2020scanrefer}, 3D-VisTA \cite{zhu20233d}, and GPS trained on SceneVerse \cite{jia2024sceneverse}. The best performing method, GPS, however, still only achieves an accuracy of 64.9\% on natural language statements \cite{jia2024sceneverse}, which is far below the acceptable threshold for real-world deployment. All of these models are also based on a similar transformer architecture, and with the exception of GPS, cannot handle open-vocabulary object names, which is unideal for a real-world use case. These models' ability to only choose the most likely object from a list also makes them incapable of handling situations where the language input has mistakes or is only partially valid.
\section{VLA-3D Dataset}
\subsection{Overview}
\begin{figure}
\centering
\includegraphics[width=0.4\textwidth]{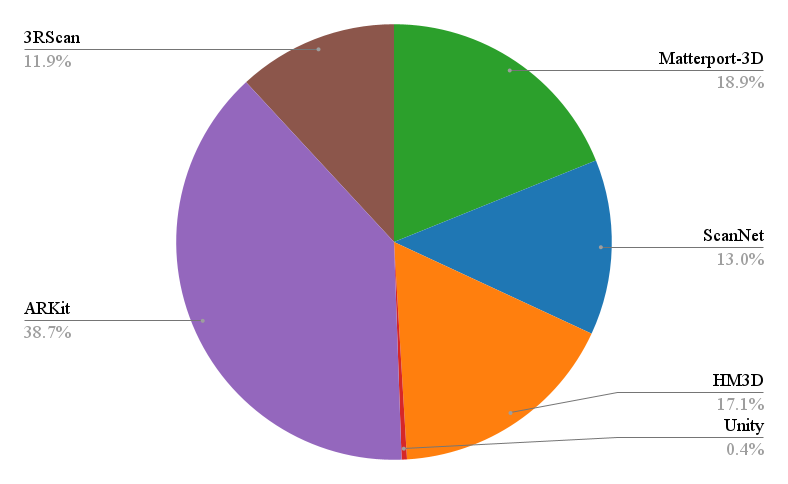}
\caption{Breakdown of regions from each data source}
\label{fig:data_sources}
\end{figure}

\begin{figure}
\centering
\includegraphics[width=0.45\textwidth]{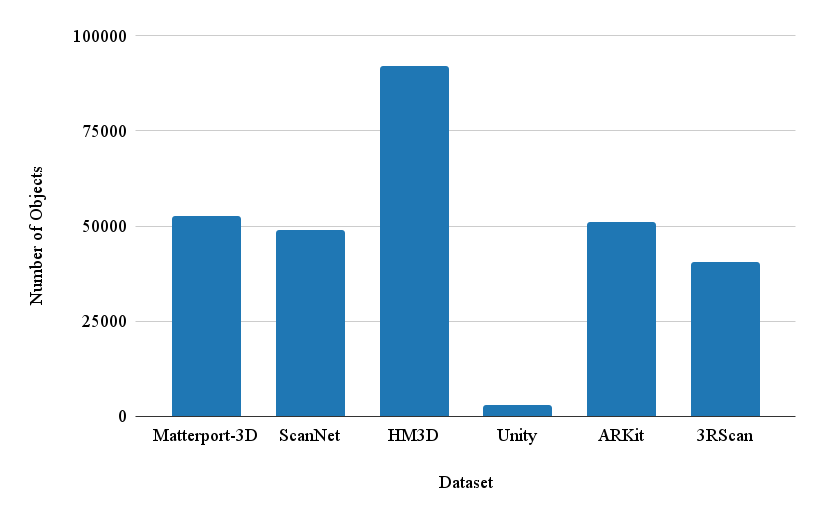}
\caption{Total number of objects in each dataset processed}
\label{fig:num_objs}
\end{figure}

\begin{figure}
\centering
\includegraphics[width=0.5\textwidth, trim={0 3cm 0 2cm}]{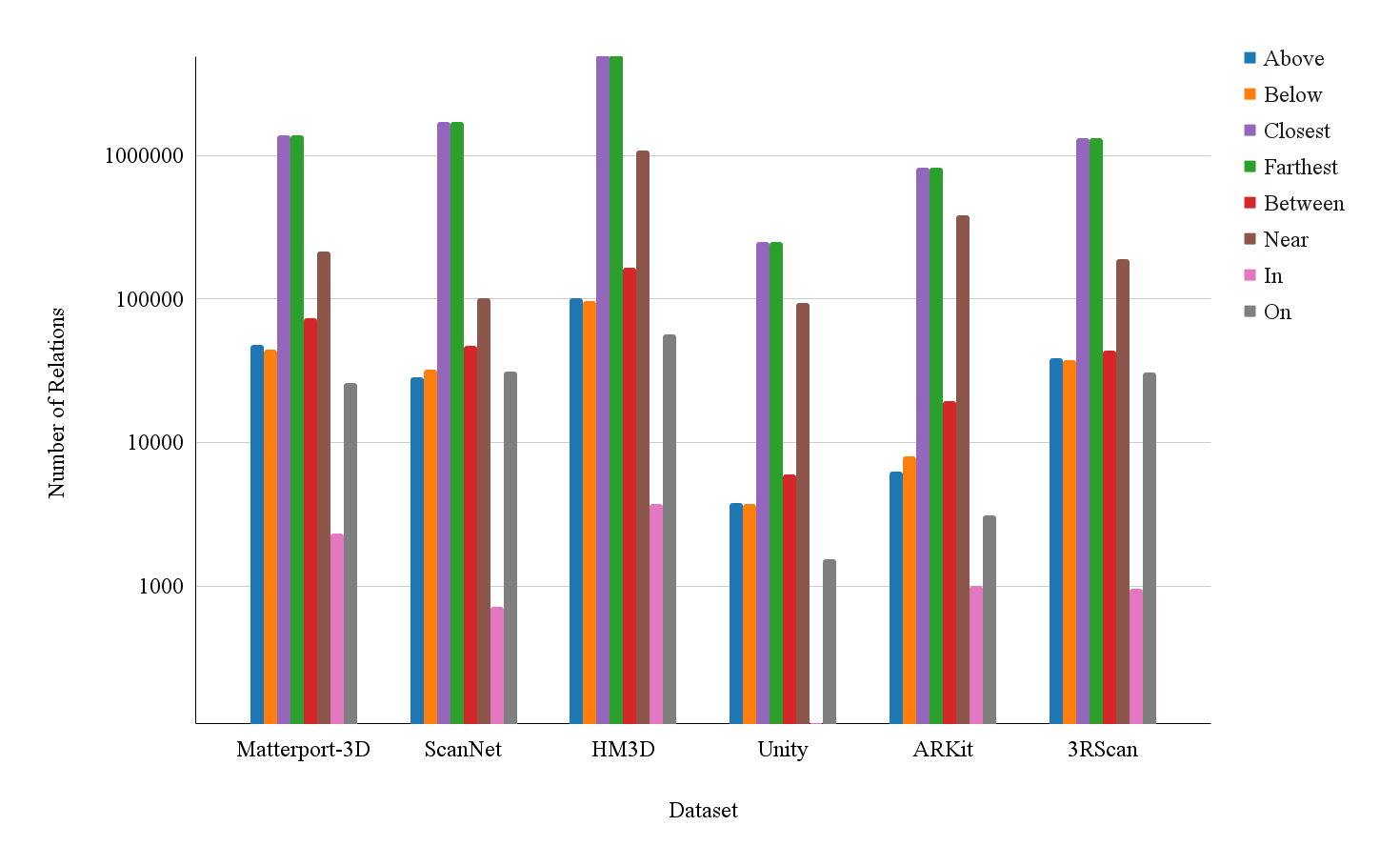}
\caption{Total number of each relation type from each dataset processed}
\label{fig:relations}
\end{figure}

To aid the development of robust and interactive indoor navigation agents, we introduce a synthetically-generated, publicly released dataset for Vision and Language-guided Action in 3D Scenes (VLA-3D). Our dataset is based on 3D scans from five real-world datasets: ScanNet \cite{dai2017scannet}, Matterport3D \cite{Matterport3D}, Habitat-Matterport 3D (HM3D) \cite{ramakrishnan2021habitat}, 3RScan \cite{Wald2019RIO}, and ARKitScenes \cite{baruch2021arkitscenes}, as well as scenes generated in Unity. Figure \ref{fig:data_sources} shows a breakdown of the number of regions from each data source.
For each scene, we provide:
\begin{itemize}
\item Scene point cloud
\item List of objects with semantic class labels, bounding box, and color(s)
\item List of traversable free spaces
\item List of regions with semantic labels and bounding boxes
\item Scene graph of spatial relations split by room
\item Language statements with ground-truth annotation
\end{itemize}
Two key features of our dataset are providing large-scale scene graphs for each scene that are robust to scene changes and enables identification of similar objects, as well as incorporating traversable free space as referential targets in addition to just objects. In total, our dataset contains 7635 scenes which contain over 11.5k regions, defined as separate rooms in a scene. A total of over 286k objects from 477 unique classes exist in the dataset, along with 23.5M inter-object spatial relations and 9.7M referential statements. Figure \ref{fig:num_objs} shows the total number of objects in each dataset source while Figure \ref{fig:relations} shows the number of each spatial relation generated per dataset.

\begin{figure*}[t!]
\centering
\includegraphics[width=\textwidth, trim={0 2.2cm 0.5cm 2cm},clip]{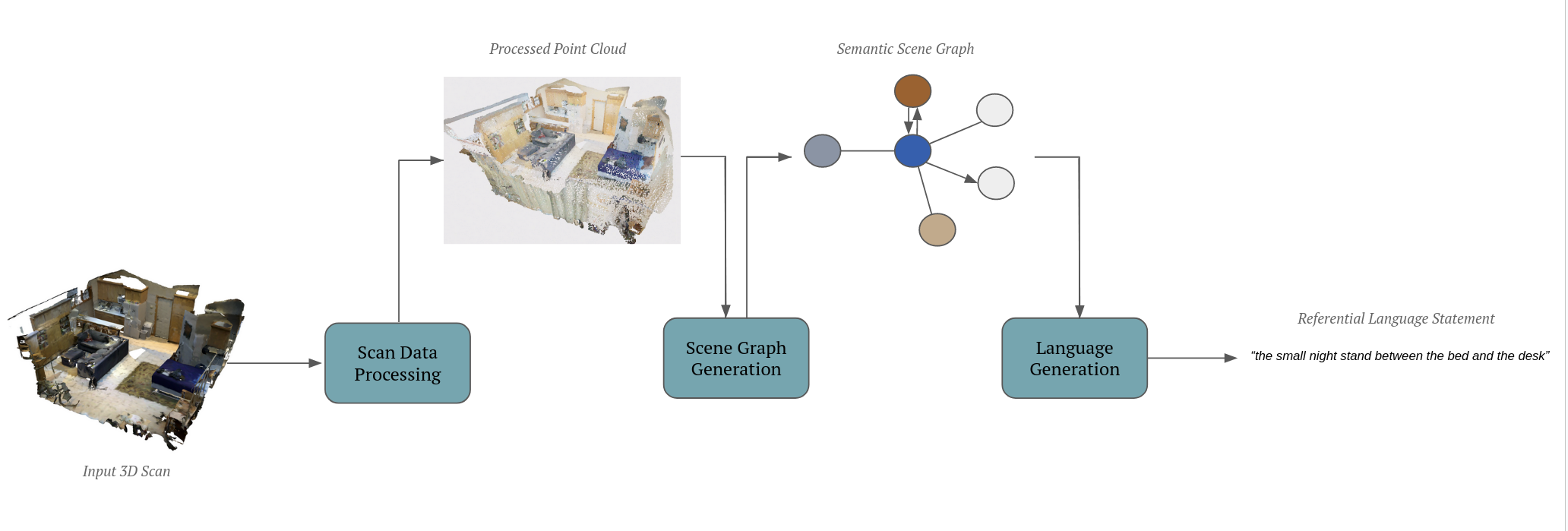}
\caption{Data processing pipeline consisting of: 3D Scan Processing, Scene Graph Generation, and Language Generation}
\label{fig:data_processing}
\end{figure*}

The data curation process is further detailed below and an overview is shown in \ref{fig:data_processing}.

\subsection{3D Scan Processing}
To generate point cloud files, scene-level point clouds were obtained from the vertices defined in the original PLY files for ScanNet, Matterport3D, and ARKitScenes. For HM3D, Unity, and 3RScan scenes, point clouds were sampled uniformly from the original mesh files while colors were sampled from the textures. Regions and objects were identified leveraging the semantic information in the original meshes. ScanNet, 3RScan, and ARKitScenes each have a single room per scene while region segmentations are provided in Matterport-3D and HM3D, and custom-segmented for Unity scenes. For each object labeled in the scenes, an open-vocabulary class name is stored and the semantic class is mapped to both the NYU40 \cite{gupta2013perceptual} and NYUv2 \cite{silberman2012indoor} schemas with the provided mappings for flexibility \footnote{For the Unity scenes, the ground-truth semantic labels were cleaned then manually mapped to the class schemas by five data annotators. A validation round was done to standardize the labels.}. The dominant three colors (if any) were obtained for each object based on the object point cloud and a color clustering algorithm.

To provide extra navigation targets, each scan was also processed to generate the horizontally traversable free space. Separate traversable regions in a room are chunked into sub-regions, for which spatial relations with other objects in the scene are generated to create unambiguous references to these spaces (e.g. ``the space near the table").

\subsection{Scene Graph Generation}
Eight different types of semantic spatial relations were calculated using heuristics based on the yawed object bounding boxes to generate a scene graph of relations. Relations are generated exhaustively for every pair or triple of objects within a region, then filtered afterwards depending on the semantic classes involved. All relations are binary except for the ``\textit{between}" relation, which is ternary. 

Table \ref{tab:relations} defines the types of spatial relations used.
\begin{table}[t!]
\caption{Summary of semantic relationship types in VLA-3D}
\label{tab:relations}
\vspace{-2em}
\begin{center}
\resizebox{\linewidth}{!}{
  \begin{tabular}{|l|p{3.5cm}|p{2.5cm}|l|}
     \hline
     \textbf{Relation} & \textbf{Definition} & \textbf{Synonyms} & \textbf{Properties} \\ 
     \hline
     Above & Target is above the anchor & Over &  \\
     \hline
     Below & Target is below the anchor & Under, Beneath, Underneath &  \\
     \hline
     Closest & Target is the closest object of a certain class to the anchor & Nearest & Inter-class \\
     \hline
     Farthest & Target is the farthest object from a certain class to the anchor & Most distant, Farthest away & Inter-class \\
     \hline
     Between & Target is between two anchors & In the middle of, In-between & Ternary \\
     \hline
     Near & Target is within a threshold distance of the anchor & Next to, Close to, Adjacent to, Beside & Symmetric \\
     \hline
     In & Target is inside the anchor & Inside, Within &  \\
     \hline
     On & Target is above and in contact with the anchor in the Z-axis & On top of &  \\
     \hline
  \end{tabular}}
\end{center}
\end{table}

\subsection{Language Generation}
Referential language statements were synthetically generated based on the computed scene graph using a template-based generation method. From the table above, synonyms for each relation are used to add variety into the statements. Every statement has at least one semantic relation and only uses object attributes if needed to distinguish the target object. The generated statements are also:
\begin{enumerate}
\item \textbf{View-independent}: The relation predicate for the target object does not depend on the perspective from which the scene is viewed from.
\item \textbf{Unambiguous}: Only one possibility exists in the region for the referred target object.
\item \textbf{Minimal}: Following Grice's maxim of manner \cite{sep-grice}, statements use the least possible descriptors to disambiguate the target object.
\end{enumerate}

\section{Baseline Evaluation}

To verify the difficulty of our dataset, we evaluate the pre-trained checkpoints of two SOTA open-source baseline models on our data directly: MVT and 3D-VisTA. MVT is the best-performing method on the official ReferIt3D benchmark while 3D-VisTA is a more recent method that has since outperformed MVT. The test results are shown in Table \ref{tab:baselines}. The test performances on both Nr3D and Sr3D (which the models are trained on) are also shown as a point of comparison.

\begin{table}[t!]
\caption{Referential Object Grounding Accuracy of Two Baseline Models on VLA-3D, Nr3D, and Sr3D}
\label{tab:baselines}
\begin{center}
  \begin{tabular}{|c|c|c|c|}
     \hline
     \textbf{Baseline Model} & \textbf{VLA-3D} & \textbf{Nr3D} & \textbf{Sr3D} \\
     \hline
      MVT & 22.5\% & 59.5\% & 64.5\% \\
     \hline
     3D-VisTA & 28.9\% & 64.2\% & 76.4\% \\
    \hline
  \end{tabular}
\end{center}
\end{table}

The results of both models are much lower on our dataset compared to their performance on the ReferIt3D benchmark, likely due to the fact that they are directly evaluated not just on new language data but also on unseen scenes with many more fine-grained objects than what they were trained on. Upon examining failure cases, we observe that failures are either due to object classification errors, language semantic reasoning errors (e.g. mixing up target and anchor object), or errors in spatial reasoning (e.g. choosing ``distractor objects" of the same semantic class but incorrect spatial relation). This disconnect in performance indicates the poor cross-domain generalizability of existing methods, especially to complex real-world scenes, and delineates the need for more diverse language data to improve 3D visual grounding models and enable their use in more complex tasks like interactive indoor navigation. It also verifies VLA-3D as a challenging benchmark for progress towards this goal.

\section{Conclusion} 
\label{sec:conclusion}
Aiming to advance progress in semantic scene reasoning and understanding in robotics applications, we introduce a large-scale novel dataset of object-referential natural language statements along with spatial scene graphs for a diverse set of 3D scenes. This dataset contains a variety of spatial relationships and language statements on the scale of millions and is suited for the sub-task of referential object grounding guided by structured scene representations. 
Future extensions to VLA-3D include augmenting the statements with LLMs, adding 3D scan data from other real-world sources, generating compound relational statements, generating view-dependent statements, and extending the statements beyond referential object-grounding to include the action component explicitly. Further research using this dataset for training could involve the development of generalizable system-integrated modules with the capabilities of answering questions about the scene, identifying items not in the scene, and suggesting alternative objects with similar attributes, location, or affordances. Overall, our dataset establishes a resource for the development of generalizable methods that extract observations from 3D scenes and reason about them using open-vocabulary natural language, which aids the development of interactive indoor navigation agents that can operate in changing environments, both alongside and with humans.

\bibliographystyle{plainnat}
\bibliography{references}

\end{document}